\documentclass[conference,compsoc]{IEEEtran}
\usepackage{amsmath}
\usepackage{amssymb}
\usepackage{float}
\usepackage[hidelinks]{hyperref}
\usepackage{subfig}                                                                                                                                           
\usepackage{tabularx}
\usepackage{tikz}

\graphicspath{{figures/}}
\usetikzlibrary{positioning}
\usetikzlibrary{shapes.geometric, arrows, backgrounds}

\begin{document}
\title{Camera-Based Adaptive Trajectory Guidance via Neural Networks}

\author{\IEEEauthorblockN{Aditya Rajguru}
\IEEEauthorblockA{Computer Science and Engineering\\
University of Texas Arlington\\
Arlington, TX, USA\\
aditya.rajguru@mavs.uta.edu}
\and
\IEEEauthorblockN{Christopher Collander}
\IEEEauthorblockA{Computer Science and Engineering\\
University of Texas Arlington\\
Arlington, TX, USA\\
christopher.collander@mavs.uta.edu}
\and
\IEEEauthorblockN{William J. Beksi}
\IEEEauthorblockA{Computer Science and Engineering\\
University of Texas Arlington\\
Arlington, TX, USA\\
william.beksi@uta.edu}}

\maketitle
\begin{abstract}
In this paper, we introduce a novel method to capture visual trajectories for
navigating an indoor robot in dynamic settings using streaming image data. First, 
an image processing pipeline is proposed to accurately segment trajectories from 
noisy backgrounds. Next, the captured trajectories are used to design, train, 
and compare two neural network architectures for predicting acceleration and 
steering commands for a line following robot over a continuous space in real 
time. Lastly, experimental results demonstrate the performance of the neural 
networks versus human teleoperation of the robot and the viability of the system 
in environments with occlusions and/or low-light conditions. 
\end{abstract}


\section{Introduction}
\label{sec:introduction}                                                                                                                                      
Line following robots have a variety of use cases in education, entertainment, 
health care, factory/warehouse settings, and more \cite{su2010intelligent,
colak2009evolving,ang2013automated,jain2014applications,scholar2016serving}. 
However, the effectiveness of these types of mobile robots in realistic 
deployments is dependent upon their ability to negotiate environments in which 
obstacles are dynamic, trajectories are occluded, and lighting conditions vary. 
Existing control methods for robot line following utilize knowledge-based 
approaches that rely upon structured environments. While effective in simplistic 
scenarios, these methods are severely limited by their surroundings. Moreover, 
they do not generalize for real-world use due to the inherent uncertainty in 
states encountered during operation. 

\begin{figure}[ht]
\centering
\includegraphics[scale=0.325]{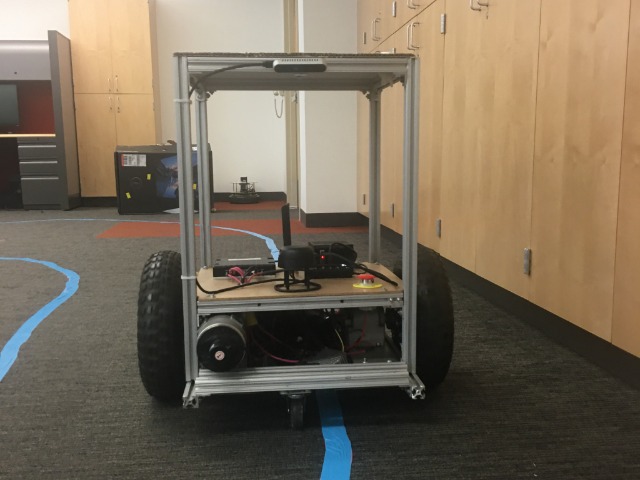}
\caption{A line following robot equipped with a stereo camera.}
\label{fig:line_following_robot}
\end{figure}

Analog sensors have historically been the most common approach to provide 
control for a line following robot. These sensors can provide inference in 
high-contrast binary environments, but often require dedicated circuits and fail 
when encountering discontinuous trajectories. Vision-based approaches such as 
sensor array matrices use the location of image pixel values to generate 
pulse-width modulation outputs for motor control. However, the assumption that 
information will be found in a certain region of interest is not always true. In 
comparison, a neural network can be trained to produce control 
outputs based on noisy images where information is contained in different areas
and trajectories are discontinuous. 

Neural network-based methods for line following often discretize the output 
space into steering commands by using classification models that only allow for 
lateral control. Although this enables the robot to steer correctly, it does not 
make use of the trajectory information to provide control longitudinally. In
addition, discretizing the output action space results in a loss of precision. A 
continuous output space over both lateral and longitudinal velocities allows for 
reactive control based on the steepness of the trajectory, hence mimicking 
human behavior.

To address these issues, we first propose a decision tree thresholding algorithm 
that can adapt to active and cluttered environments using a minimal amount of 
surveyed data. Next, we design and compare the performance of two neural network 
architectures: a multilayer perceptron (MLP) and a 1D convolutional neural 
network (CNN). Our networks produce continuous output over both linear and 
angular velocities using a regression-based model. Finally, we demonstrate the 
usability of the system through the comparison of decisions made by the neural 
networks against those of a human operator with a line following robot, Figure 
\ref{fig:line_following_robot}.

The remainder of this paper is organized as follows. Related research work is
reviewed in Section \ref{sec:related_work}. Section \ref{sec:approach} provides 
the details of our proposed approach. Experimental results are presented and
discussed in Section \ref{sec:experimental_results}. We conclude in Section 
\ref{sec:conclusion}.


\section{Related Work}
\label{sec:related_work}                                                                                                                                      
Line following robots have traditionally used comparator circuits with analog 
sensors to detect the presence of a trajectory. For example, Punetha et al. 
implement an array of light dependent resistors and IR proximity sensors to 
follow a line in a binary environment with sharp contrast differences
\cite{punetha2013development}. Their approach works in controlled environments,
yet real-world conditions usually do not offer such simple scenarios and 
trajectories may not always be continuous. 

Ismail et al. take a vision-based approach to line following 
\cite{ismail2009vision}. The authors generate proportional pulse-width modulated 
outputs directly from images using a score calculated from a sensor array matrix 
under the assumption that pixels lie in the center of the image. However, if 
there are pixels denoting a straight line towards the periphery of the image the 
system will fail to predict accurate control outputs. Similarly, Rahman et al. 
propose a line following vision system assuming that objects in the environment 
are static \cite{rahman2005architecture}. Although this approach may work in 
constrained environments, the expectation of a static environment contradicts 
most real world scenarios. 

Pomerleau uses a camera feed and laser rangefinder measurements with a 
single-layer neural network to a steer vehicle \cite{pomerleau1989alvinn}. The 
activations of the network's input layer are proportional to the blue channel of 
the image. Nevertheless, the activations of the network are not dependent on 
color since the implementation discards this information before passing the 
input to the network. In another implementation using a CNN for line following, 
Borne and Lowrance discretize the output space to produce steering commands 
\cite{born2018application}. Likewise, Tai et al. implement a CNN with 
fully connected end layers activated using a softmax function to predict 
probabilities of members in a discrete steering action space multiplied with 
preset velocities \cite{tai2016deep}. 


\section{Approach}
\label{sec:approach}                                                                                                                                      
In this section, we describe our approach to obtain visual trajectories in 
dynamic environments from a streaming camera feed. We use these trajectories 
with two different neural networks architectures to provide acceleration and 
steering control for a mobile robot over a continuous space in real time. The 
performance of the networks is compared to human teleoperation of a line 
following robot. 

\subsection{Image Preprocessing}
\begin{figure}[ht]
\centering
\includegraphics[scale=0.25]{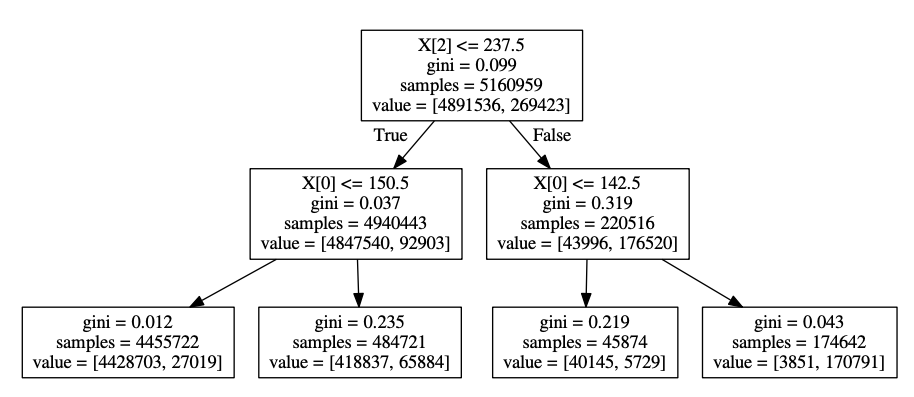}
\caption{The decision tree used for colored line segmentation.}
\label{fig:decision_tree}
\end{figure}

To accurately segment the trajectories from noisy backgrounds, obstacles that 
move even after the field of view is restricted, and be tolerant to fluctuating 
lighting conditions, a HSV (hue, saturation, and value) threshold was trained. 
Since the robot may follow different colored lines, individual HSV thresholds 
must be learned. First, the robot's operational environment was surveyed and 
approximately 20 images were collected for a single color of track in diverse 
conditions (e.g. under various lighting conditions, backgrounds, objects in the 
frame, etc.).

After collecting the data, each pixel was labeled as either line or non-line. 
The labeled data was then passed to decision tree (DT) classifier using the CART 
algorithm \cite{denison1998bayesian} with a Gini impurity measure for making the 
splits. The maximum depth level of the DT was 2. This resulted in a 98\% mean 
accuracy and the ability to threshold the image with conditionals on the hue and 
value parameters of the HSV color space, Figure \ref{fig:decision_tree}. 
Separate DTs were trained for different colors and their values are chosen 
according to the user input. 

\begin{figure}[ht]
\tikzstyle{startstop} = [rectangle, rounded corners, minimum width=1cm, minimum height=1cm,text centered, draw=black, fill=red!30,scale=0.5]
\tikzstyle{io} = [trapezium, trapezium left angle=70, trapezium right angle=110, minimum width=3cm, minimum height=1cm, text centered, draw=black, fill=blue!30,scale=0.5]
\tikzstyle{process} = [rectangle, minimum width=1cm, minimum height=1cm, text centered,text width =2cm, draw=black, fill=orange!30,scale=0.5]
\tikzstyle{decision} = [diamond, minimum width=3cm, minimum height=1cm, text centered, draw=black, fill=green!30,scale=0.5]
\tikzstyle{arrow} = [thick,->,>=stealth]
\begin{tikzpicture}[node distance=3cm]
 \node (in1) [startstop] {Raw RGB Image};
 \node (pro1) [process, right of =in1] {Gaussian Blur};
 \node (pro2) [process, right of = pro1] {HSV Conversion};
 \node (pro3) [process, right of = pro2] {Thresholding with DT};
 \node (pro4) [process, right of = pro3] {Resize ($32\times32$)};
 \node (pro5) [process, right of = pro4] {Image Binarization};
    
 \draw [arrow] (in1) -- (pro1);
 \draw [arrow] (pro1) -- (pro2);
 \draw [arrow] (pro2) -- (pro3);
 \draw [arrow] (pro3) -- (pro4);
 \draw [arrow] (pro4) -- (pro5);
\end{tikzpicture}
\caption{The image preprocessing pipeline for robot line following.}
\label{fig:image_preprocessing_pipeline}
\end{figure}
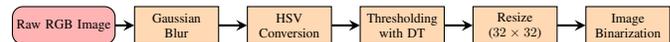
    
All the images are initially filtered through an image preprocessing pipeline 
before being given to the neural networks, Figure 
\ref{fig:image_preprocessing_pipeline}. First, a Gaussian blur is applied to 
smooth the image. Then, the RGB image is converted to the HSV color space. After 
the conversion, a threshold specific to the line color that was learned by the 
DT is applied and a segmented image of the trajectory is obtained. The segmented
trajectory is downsampled to $32 \times 32$ pixels to reduce computational 
complexity. Afterwards, the images are binarized and subsequently flattened for 
input to the neural networks.

\subsection{Trajectory Data Collection}
We take a learning by demonstration approach to train the neural networks 
\cite{argall2009survey}. A track was laid out in a lighting controlled 
environment and the dataset was manually collected. Multiple rounds of data 
collection were performed, each with a varied camera orientation, frame rate,
and track color. The robot was teleoperated from a base station with the 
operator judging movement purely on the incoming images. The processed images 
and velocity outputs were recorded to a CSV file. To avoid unintentional 
learning of user speeds, the velocities were normalized to a unit vector. 

\begin{figure}[ht]
\centering
\subfloat[Angular velocities.] {
  \includegraphics[scale=0.25]{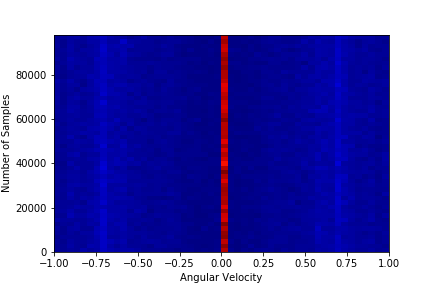}
}
\subfloat[Linear velocities.] {
  \includegraphics[scale=0.25]{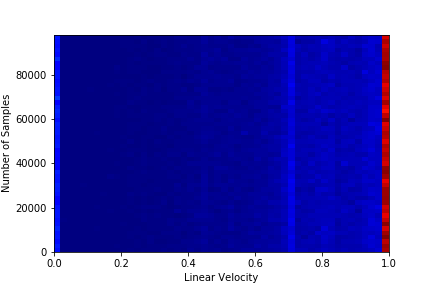}
}
\caption{The trajectory dataset distribution heatmaps.}
\label{fig:distribution_heatmaps}
\end{figure}

The dataset was augmented by mirroring the images and negating the corresponding 
angular velocity. The final dataset consisted of 122,576 labeled images with a 
72/20/8 training/test/validation split. Note that the distribution of velocities 
in the dataset is not uniformly distributed as shown by the heatmaps in Figure 
\ref{fig:distribution_heatmaps}.

\subsection{Neural Networks}
In this subsection, we propose two neural network architectures for predicting
the linear and angular steering velocities of a line following robot. A detailed 
justification is provided for the choice of the networks along with the network 
hyperparameters. In addition, we compare the validation versus training loss for 
each network. 


\begin{table}[ht]
\begin{tabularx}{0.45\textwidth} { 
  | >{\raggedright\arraybackslash}X 
  || >{\centering\arraybackslash}X 
  | >{\raggedleft\arraybackslash}X |}
\hline
\textbf{Layer Type} & \textbf{Hyperparameters} & \textbf{Output Shape} \\
\hline
Input  & 1024 & (1,1024) \\
\hline
Convolution1D (Filter Size - 3) & 307 & (307,1022) \\
\hline
Dropout & 20\% Dropout & (307,1022) \\
\hline
Max Pooling & Pool Size - 3 & (102,1022) \\
\hline
Dense & 207 & (102,207)\\
\hline
Batch Normalization & None & (102,207)\\
\hline
Dropout & 10\% Dropout & (102,207)\\
\hline
Convolution1D (Filter Size -1) & 100 & (102,100)\\
\hline
Dense & 100 & (102,100)\\
\hline
Batch Normalization & None & (102,100)\\
\hline
Dropout & 20\% Dropout & (102,100)\\
\hline
Dropout & 20\% Dropout & (102,100)\\
\hline
Flatten & None & (1,10200)\\
\hline
Dense & 2 & (1,2)\\
\hline
\end{tabularx}
\caption{1D CNN architecture.}
\label{tab:1d_cnn_architecture}
\end{table}

\subsubsection{1D CNN}
2D CNNs are commonly used to detect local features in images. By using an 
overcomplete set of filters, variations of patterns can be learned for specific 
local features and can therefore produce accurate local feature maps. Higher 
level feature maps in 2D CNNs correspond to larger input region areas. Thus,
generating abstractions by combining the lower level features can lead to good 
performance in tasks where spatial relationships are important 
\cite{lin2013network}. In this work, most spatial relationships are not 
significant since the actions depend more on global features rather than local 
features. Despite many architectural models tested, a 2D CNN failed to converge 
for our line following application. 1D CNNs have performed well in signal and 
image processing applications \cite{1dcnnsignal}. Due to the sparsity of our 
data, a 1D CNN architecture was implemented with several fully connected layers 
for regression over the output velocities, Table
\ref{tab:1d_cnn_architecture}.

\begin{figure}[ht]
\centering
\includegraphics[scale=0.35]{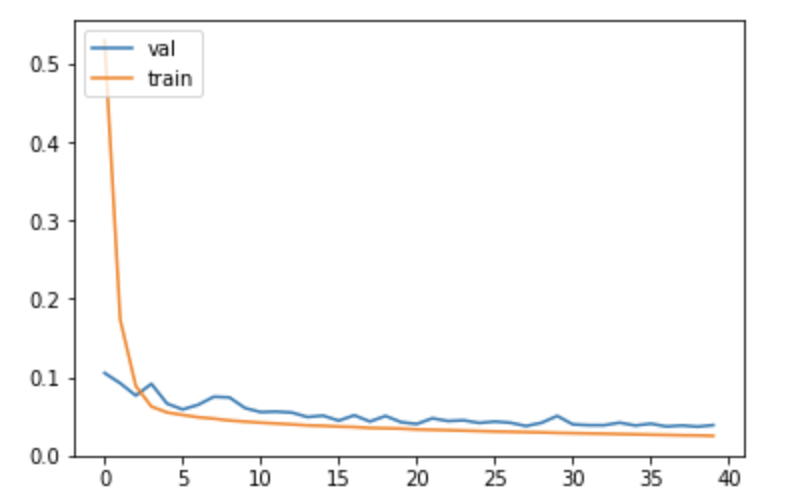}
\caption{Validation versus training loss using a 1D CNN.}
\label{fig:1d_cnn_loss}
\end{figure}

The preprocessed input image was flattened and fed into a 1D convolution layer 
which generated 307 feature maps. The activation used after each convolution
layer is the softsign function,
\begin{equation*}
  y = \frac{x}{1+\left | x \right |}.
\end{equation*}
In the final layer the activation is linear. The Adam \cite{kingma2014adam} 
optimizer is used with a learning rate of 0.0001, an exponential decay rate of 
0.9, and a mean squared error loss function. Max pooling, dropouts 
\cite{srivastava2014dropout} and batch normalization \cite{ioffe2015batch} are 
regularly used throughout the network to generalize and prevent overfitting. 
The validation and training loss graphs for the network trained on the datasets 
are presented in Figure \ref{fig:1d_cnn_loss}.

\subsubsection{Multilayer Perceptron}
\begin{table}[ht]
\begin{tabularx}{0.45\textwidth} { 
  | >{\raggedright\arraybackslash}X 
  || >{\centering\arraybackslash}X 
  | >{\raggedleft\arraybackslash}X | }
 \hline
\textbf{Layer Type} & \textbf{Hyperparameters} & \textbf{Input Shape}  \\
 \hline
 Input  & 1024 & (1,1024) \\
\hline
Dense-1 & 300 & (1024,300) \\
\hline
Dropout & 20\% Dropout & (1024,300) \\
\hline
Dense-2 & 200 & (300,200) \\
\hline
BatchNorm & None & (300,200)\\
\hline
Dropout & 10\% Dropout & (300,200)\\
\hline
Dense-3 & 200 & (300,200)\\
\hline
Output & 2 & (200,2)\\
\hline
\end{tabularx}
\caption{The MLP architecture.}
\label{tab:mlp_architecture}
\end{table}

\tikzset{
  every neuron/.style={
    circle,
    draw,
    minimum size=1cm
  },
  neuron missing/.style={
    draw=none, 
    scale=4,
    text height=0.333cm,
    execute at begin node=\color{black}$\vdots$
  },
}

The MLP architecture consists of three fully connected hidden layers and a
varying number of nodes, Table \ref{tab:mlp_architecture}. The input layer 
consists of 1,024 nodes. Each node corresponds to a pixel of the preprocessed 
image and is forward propagated to produce two outputs at the last layer. These
outputs correspond to the angular and linear velocities. Due to additional 
parameters and the suitability of the MLP for regression-based problems 
\cite{DBLP:journals/corr/abs-1810-11333}, the accuracy metrics and performance 
achieved was slightly better than the 1D CNN model.

\begin{figure}[ht]
\centering
\includegraphics[scale=0.35]{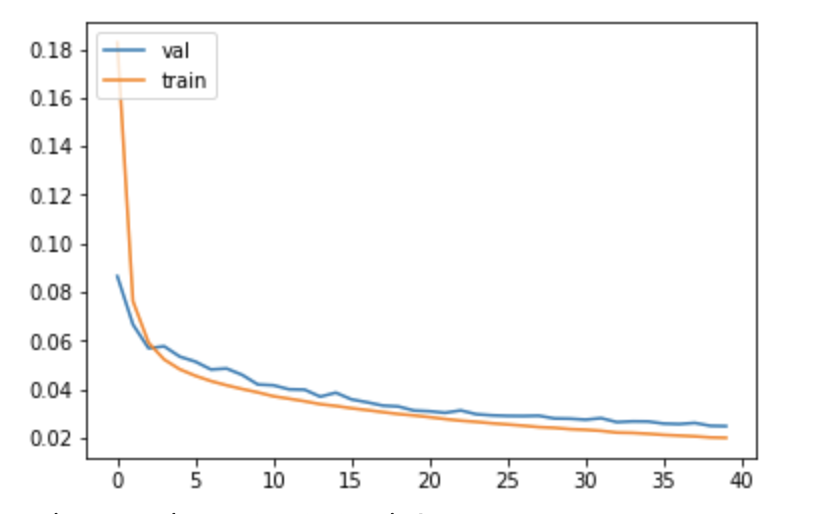}
\caption{Validation versus training loss using a MLP.}
\label{fig:mlp_loss}
\end{figure}

For the activation functions, the ReLU function was used for all the layers
except for the output layer. The optimizer and the loss function are identical 
to the ones used in the 1D CNN architecture. Due to the larger dimensionality of 
the MLP, the network is prone to overfitting. Dropout and batch normalization 
were used in the network to help alleviate these issues. To check for 
overfitting, the training and validation loss were calculated and observed 
throughout training. Due to the absence of an inflection point and the presence 
of a steady decrease in both validation and training losses, overfitting was 
minimized as shown in Figure \ref{fig:mlp_loss}.

\section{Experimental Results}
\label{sec:experimental_results}
In this section, we present the experimental results of our methods. The
experiments were conducted at the University of Texas at Arlington Robotic
Vision Laboratory. All experiments were conducted on a line following robot 
under realistic and challenging environmental conditions. 

\subsection{Robot Description}
A robotic research platform was used for data collection, training, and 
experimental testing of the camera-based adaptive trajectory guidance system 
(Figure \ref{fig:line_following_robot}). The robot consists 
of a custom built 1.5 ft $\times$ 1.5 ft chassis and is powered by two 
wheelchair motors using a differential drive configuration through a PID enabled 
Roboteq motor controller. The on-board computer is a Nivdia Jetson AGX Xavier 
running the Robot Operating System (ROS) \cite{quigley2009ros}. ROS is used to 
communicate between the various systems of the robot, including the Intel 
RealSense D415 stereo camera. The camera has a resolution of $640 \times 480$ 
and a frame rate set to 6 fps. In addition, the camera was angled downward to 
provide a better view of the ground.

\subsection{Model Metrics}
\begin{table}
\begin{tabular}{l|l|l|l|l|}
\cline{2-5}
                               & \multicolumn{2}{c|}{\textbf{1D CNN}} & \multicolumn{2}{c|}{\textbf{MLP}} \\ \cline{2-5} 
                               & Train & Validation & Train & Validation \\ \hline
\multicolumn{1}{|l|}{Accuracy} & 0.924 & 0.884      & 0.925 & 0.907\\ \hline
\multicolumn{1}{|l|}{Loss}     & 0.025 & 0.038      & 0.020 & 0.025\\ \hline
\end{tabular}
\caption{The accuracy and loss for the neural networks on the training and
validation datasets.}
\label{tab:accuracy_loss}
\end{table}

\begin{table}
\begin{tabular}{l|l|l|l|l|}
\cline{2-5}
                           & \multicolumn{2}{c|}{\textbf{1D CNN}} & \multicolumn{2}{c|}{\textbf{MLP}} \\ \cline{2-5} 
                           & Linear       & Angular       & Linear     & Angular     \\ \hline
\multicolumn{1}{|l|}{MAE}  & 0.010        & 0.183         & 0.090      & 0.162       \\ \hline
\multicolumn{1}{|l|}{RMSE} & 0.148        & 0.258         & 0.142      & 0.247       \\ \hline
\end{tabular}
\caption{The mean absolute error (MAE) and root mean square error (RMSE) for the
neural networks.}
\label{tab:mae_rmse}
\end{table}

The model metrics obtained while training are displayed in Tables 
\ref{tab:accuracy_loss} and \ref{tab:mae_rmse}. The MLP has a lower validation 
loss and higher validation accuracy than the 1D CNN. To test the performance of 
the model, the root mean square error and mean absolute error were calculated on 
24,516 test samples. The 1D CNN performed as good as the MLP using 35\% less 
parameters (265,519 vs. 409,102), while the performance of both models was 
similar.

\subsection{Human Operator Versus Neural Network Decision Comparison}
\begin{figure}[ht]
\centering
\subfloat[MLP linear performance.] {      
  \includegraphics[scale=0.25]{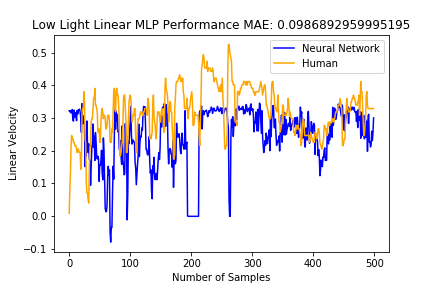}
}
\subfloat[MLP angular performance.] {      
  \includegraphics[scale=0.25]{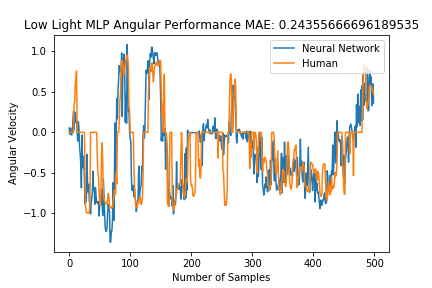}
}\\
\subfloat[CNN linear performance.] {      
  \includegraphics[scale=0.25]{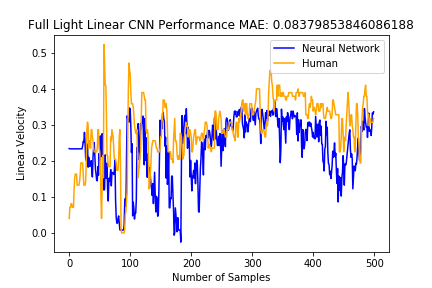}
}
\subfloat[CNN angular performance.] {      
  \includegraphics[scale=0.25]{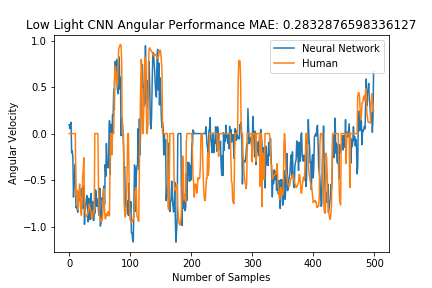}
}
\caption{The velocity decisions of the human operator versus the neural network.}
\label{fig:human_vs_neural_network}
\end{figure}

To test the viability of our system in changing conditions, a comparison between 
human teleoperation velocities and neural network predicted velocities was 
performed (Figure \ref{fig:human_vs_neural_network}). Both the 1D CNN and the 
MLP architectures were tested against a new test track with a different color. 
To enable the neural network predictions, a threshold was learned for the new 
color and the image processing pipeline was updated accordingly. It is worth 
noting that the mean absolute error differs with each teleoperation as no two 
human operated runs will be exactly the same. Although the mean absolute error 
is a bit higher than the test values, even under low-light conditions the 
performance does not degrade much and it follows the human decision with a 
moderate amount of error (Table \ref{tab:human_vs_neural_network}). We also 
observe from the 1D CNN and MLP linear performance graphs that the robot is able 
to adjust its linear velocity based on the trajectory ahead similar to a human 
operator. To portray the information clearly, the linear velocity is scaled down
and has not been normalized. 

\begin{table}[ht]
\begin{tabularx}{0.5\textwidth} { 
  | >{\raggedright\arraybackslash}X 
  || >{\centering\arraybackslash}X 
  | >{\raggedleft\arraybackslash}X | }
 \hline
\textbf{Model} & \textbf{Angular MAE} & \textbf{Linear MAE}  \\
 \hline
 MLP (Low Light)    & 0.244 & 0.177 \\
 \hline
 CNN (Low Light)    & 0.283 & 0.259 \\
  \hline
 MLP (Full Light) & 0.266 & 0.163 \\
 \hline
 CNN (Full Light) & 0.333 & 0.196 \\
 \hline
\end{tabularx}
\caption{The mean absolute error in human versus neural network decision.}
\label{tab:human_vs_neural_network}
\end{table}

\subsection{Occlusion Scenarios}
\begin{figure}[ht]
\centering
\includegraphics[width=0.45\textwidth]{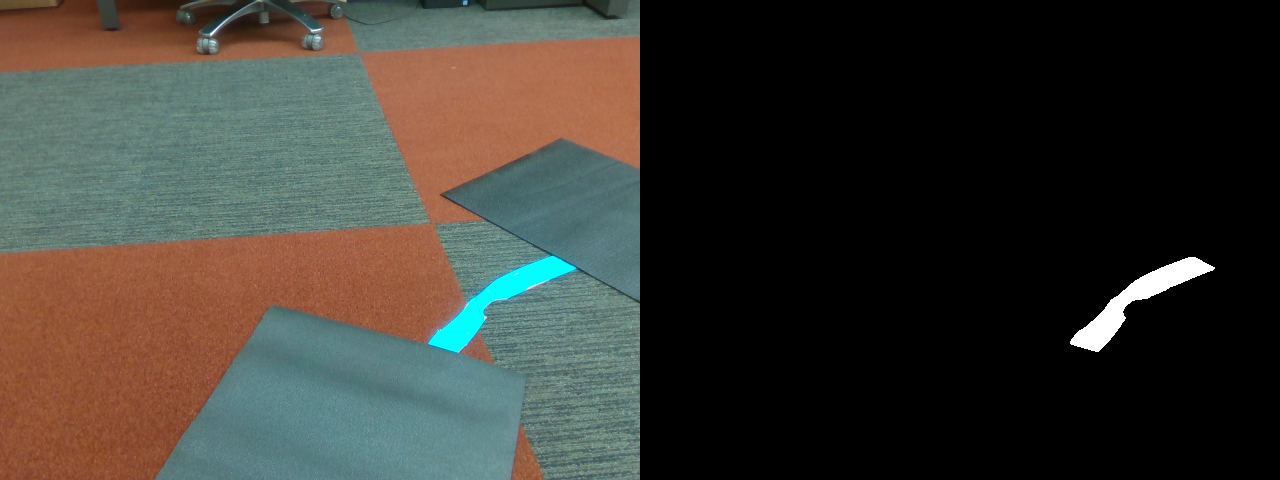}
\caption{An example a trajectory produced (right) by an occluded track (left).}
\label{fig:occuluded_trajectory}
\end{figure}

In this experiment, we test how well the robot performs when the track is 
occluded. To do this, multiple sections of the track were blocked to show 
only small portions of the curved trajectory and the predicted velocities were 
recorded (Figure \ref{fig:occuluded_trajectory}). In these tests, the robot 
correctly predicts the movement under the condition that at least one section of 
trajectory is consistently present in each camera frame. 

\section{Conclusion}
\label{sec:conclusion}                                                                                                                                      
In summary, we have presented a system for adapting a line following robot to a 
noisy, dynamic, and non-binary environment using two occlusion and lighting 
tolerant neural network architectures. The 1D CNN model achieved a performance 
close to the MLP model in all metrics, and used 35\% less parameters. In
addition, the performance of both architectures in low-light conditions as well 
as the adjustment of linear velocities was demonstrated through the comparison 
of human actions with network predictions. The experimental results showed that 
the networks perform similar to a remote human operator.

Our system can be expanded to work with more track colors through the training 
of multiple threshold DTs. For example, multiple colors of track could be used 
simultaneously to allow for live switching of thresholds based on the most 
prevalent color. Additionally, individual neural network models could be trained 
for each color possibly negating the use of a separately trained threshold. The
performance of the multi-model system may then be compared to that of a single 
model trained on the whole multi-color dataset to determine if dynamic switching 
truly provides a benefit. 

\bibliographystyle{IEEEtran}
\bibliography{camera-based_adaptive_trajectory_guidance_via_neural_networks}

\begin{thebibliography}{10}
\providecommand{\url}[1]{#1}
\csname url@samestyle\endcsname
\providecommand{\newblock}{\relax}
\providecommand{\bibinfo}[2]{#2}
\providecommand{\BIBentrySTDinterwordspacing}{\spaceskip=0pt\relax}
\providecommand{\BIBentryALTinterwordstretchfactor}{4}
\providecommand{\BIBentryALTinterwordspacing}{\spaceskip=\fontdimen2\font plus
\BIBentryALTinterwordstretchfactor\fontdimen3\font minus
  \fontdimen4\font\relax}
\providecommand{\BIBforeignlanguage}[2]{{%
\expandafter\ifx\csname l@#1\endcsname\relax
\typeout{** WARNING: IEEEtran.bst: No hyphenation pattern has been}%
\typeout{** loaded for the language `#1'. Using the pattern for}%
\typeout{** the default language instead.}%
\else
\language=\csname l@#1\endcsname
\fi
#2}}
\providecommand{\BIBdecl}{\relax}
\BIBdecl

\bibitem{su2010intelligent}
J.-H. Su, C.-S. Lee, H.-H. Huang, S.-H. Chuang, and C.-Y. Lin, ``An intelligent
  line-following robot project for introductory robot courses,'' \emph{Science
  World Transactions on Engineering and Technology Education, Lunghwa
  University of Science and Technology, Taoyuan County, Taiwan}, vol.~8, no.~4,
  pp. 1--7, 2010.

\bibitem{colak2009evolving}
I.~Colak and D.~Yildirim, ``Evolving a line following robot to use in shopping
  centers for entertainment,'' in \emph{2009 35th Annual Conference of IEEE
  Industrial Electronics}.\hskip 1em plus 0.5em minus 0.4em\relax IEEE, 2009,
  pp. 3803--3807.

\bibitem{ang2013automated}
F.~Ang, M.~K. A.~R. Gabriel, J.~Sy, J.~J.~O. Tan, and A.~C. Abad, ``Automated
  waste sorter with mobile robot delivery waste system,'' in \emph{De La Salle
  University Research Congress}, 2013, pp. 7--9.

\bibitem{jain2014applications}
T.~Jain, R.~Sharma, and S.~Chauhan, ``Applications of line follower robot in
  medical field,'' \emph{International Journal of Research}, vol.~1, no.~11,
  pp. 409--412, 2014.

\bibitem{scholar2016serving}
U.~Scholar, ``Serving robot: New generation electronic waiter,''
  \emph{International Journal of Engineering Science}, vol. 3763, 2016.

\bibitem{punetha2013development}
D.~Punetha, N.~Kumar, and V.~Mehta, ``Development and applications of line
  following robot based health care management system,'' \emph{International
  Journal of Advanced Research in Computer Engineering \& Technology
  (IJARCET)}, vol.~2, no.~8, pp. 2446--2450, 2013.

\bibitem{ismail2009vision}
A.~H. Ismail, H.~R. Ramli, M.~Ahmad, and M.~H. Marhaban, ``Vision-based system
  for line following mobile robot,'' in \emph{2009 IEEE Symposium on Industrial
  Electronics \& Applications}, vol.~2.\hskip 1em plus 0.5em minus 0.4em\relax
  IEEE, 2009, pp. 642--645.

\bibitem{rahman2005architecture}
M.~Rahman, M.~H.~R. Rahman, A.~L. Haque, and M.~T. Islam, ``Architecture of the
  vision system of a line following mobile robot operating in static
  environment,'' in \emph{2005 Pakistan Section Multitopic Conference}.\hskip
  1em plus 0.5em minus 0.4em\relax IEEE, 2005, pp. 1--8.

\bibitem{pomerleau1989alvinn}
D.~A. Pomerleau, ``Alvinn: An autonomous land vehicle in a neural network,'' in
  \emph{Advances in neural information processing systems}, 1989, pp. 305--313.

\bibitem{born2018application}
W.~Born and C.~J. Lowrance, ``Application of convolutional neural network image
  classification for a path-following robot,'' 2018.

\bibitem{tai2016deep}
L.~Tai, S.~Li, and M.~Liu, ``A deep-network solution towards model-less
  obstacle avoidance,'' in \emph{2016 IEEE/RSJ international conference on
  intelligent robots and systems (IROS)}.\hskip 1em plus 0.5em minus
  0.4em\relax IEEE, 2016, pp. 2759--2764.

\bibitem{denison1998bayesian}
D.~G. Denison, B.~K. Mallick, and A.~F. Smith, ``A bayesian cart algorithm,''
  \emph{Biometrika}, vol.~85, no.~2, pp. 363--377, 1998.

\bibitem{argall2009survey}
B.~D. Argall, S.~Chernova, M.~Veloso, and B.~Browning, ``A survey of robot
  learning from demonstration,'' \emph{Robotics and autonomous systems},
  vol.~57, no.~5, pp. 469--483, 2009.

\bibitem{lin2013network}
M.~Lin, Q.~Chen, and S.~Yan, ``Network in network,'' \emph{arXiv preprint
  arXiv:1312.4400}, 2013.

\bibitem{1dcnnsignal}
S.~{Kiranyaz}, T.~{Ince}, O.~{Abdeljaber}, O.~{Avci}, and M.~{Gabbouj}, ``1-d
  convolutional neural networks for signal processing applications,'' in
  \emph{ICASSP 2019 - 2019 IEEE International Conference on Acoustics, Speech
  and Signal Processing (ICASSP)}, May 2019, pp. 8360--8364.

\bibitem{kingma2014adam}
D.~P. Kingma and J.~Ba, ``Adam: A method for stochastic optimization,''
  \emph{arXiv preprint arXiv:1412.6980}, 2014.

\bibitem{srivastava2014dropout}
N.~Srivastava, G.~Hinton, A.~Krizhevsky, I.~Sutskever, and R.~Salakhutdinov,
  ``Dropout: a simple way to prevent neural networks from overfitting,''
  \emph{The journal of machine learning research}, vol.~15, no.~1, pp.
  1929--1958, 2014.

\bibitem{ioffe2015batch}
S.~Ioffe and C.~Szegedy, ``Batch normalization: Accelerating deep network
  training by reducing internal covariate shift,'' 2015.

\bibitem{DBLP:journals/corr/abs-1810-11333}
\BIBentryALTinterwordspacing
R.~Jankovic and A.~Amelio, ``Comparing multilayer perceptron and multiple
  regression models for predicting energy use in the balkans,'' \emph{CoRR},
  vol. abs/1810.11333, 2018. [Online]. Available:
  \url{http://arxiv.org/abs/1810.11333}
\BIBentrySTDinterwordspacing

\bibitem{quigley2009ros}
M.~Quigley, K.~Conley, B.~Gerkey, J.~Faust, T.~Foote, J.~Leibs, R.~Wheeler, and
  A.~Y. Ng, ``Ros: an open-source robot operating system,'' in \emph{ICRA
  workshop on open source software}, vol.~3, no. 3.2.\hskip 1em plus 0.5em
  minus 0.4em\relax Kobe, Japan, 2009, p.~5.

\end{thebibliography}
\end{document}